\documentclass[acmsmall]{acmart}
\usepackage{color}
\definecolor{ForestGreen}{RGB}{34,139,34}
\AtBeginDocument{%
  \providecommand\BibTeX{{%
    \normalfont B\kern-0.5em{\scshape i\kern-0.25em b}\kern-0.8em\TeX}}}

\setcopyright{acmcopyright}
\acmJournal{JETC}
\acmYear{2021} \acmVolume{1} \acmNumber{1} \acmArticle{1} \acmMonth{1} \acmPrice{15.00}\acmDOI{10.1145/3451211}

\acmPrice{15.00}
\acmISBN{978-1-4503-XXXX-X/18/06}

\begin{document}

\title{Quantization of Deep Neural Networks for
Accurate Edge Computing}

\author{Wentao Chen}
\authornote{Both authors contributed equally to this research. Wentao Chen works at Easylink Technology Co., Ltd, and this work is done when he was an visiting scholar at Guangdong Provincial People's Hospital.}
\author{Hailong Qiu}
\authornotemark[1]
\author{Jian Zhuang}
\affiliation{%
  \institution{Guangdong Cardiovascular Institute, Guangdong Provincial Key Laboratory of South China Structural Heart Disease, Guangdong Provincial People's Hospital, Guangdong Academy of Medical Sciences}
  \streetaddress{106 Zhongshan Second Road}
  \city{Guangzhou}
  \state{Guangdong}
  \postcode{510080}
}

\email{chenwentaokl@gmail.com}
\author{Chutong Zhang}
\authornotemark[1]
\author{Yu Hu}
\affiliation{%
  \institution{Easylink Technology Co., Ltd}
  %\streetaddress{P.O. Box 1212}
  \city{Wuhan}
  \state{Hubei}
  \postcode{43000}
}

\author{Qing Lu}
\author{Tianchen Wang}
\author{Yiyu Shi}
\authornotemark[2]
%\authornote{Corresponding authors.}
\email{yshi4@nd.edu}
\affiliation{%
  \institution{Department of Computer Science and Engineering, University of Notre Dame}
  %\streetaddress{P.O. Box 1212}
  \city{Notre Dame}
  \state{IN}
  \postcode{46556}
}

\author{Meiping Huang}
\authornotemark[2]
\author{Xiaowe Xu}
%\authornotemark[2]
\authornote{Corresponding authors.}
\affiliation{%
  \institution{Guangdong Cardiovascular Institute, Guangdong Provincial Key Laboratory of South China Structural Heart Disease, Guangdong Provincial People's Hospital, Guangdong Academy of Medical Sciences}
  \streetaddress{106 Zhongshan Second Road}
  \city{Guangzhou}
  \state{Guangdong}
  \postcode{510080}
}
\email{xiao.wei.xu@foxmail.com}
\orcid{0000-0002-1046-6379}

\renewcommand{\shortauthors}{W. Chen and C. Zhang, et al.}

\begin{abstract}
Deep neural networks (DNNs) have demonstrated their great potential in recent years, exceeding the performance of human experts in a wide range of applications. Due to their large sizes, however, compression techniques such as weight quantization and pruning are usually applied before they can be  accommodated on the edge. It is generally believed that quantization leads to performance degradation, and 
plenty of existing works have explored quantization strategies aiming at minimum accuracy loss. In this paper, we argue that quantization, which essentially imposes regularization on weight representations, can sometimes help to improve accuracy. 
We conduct comprehensive experiments on three widely used applications: fully connected network (FCN) for biomedical image segmentation, convolutional neural network (CNN) for image classification on ImageNet, and recurrent neural network (RNN) for automatic speech recognition, and
experimental results show that quantization can improve the accuracy by 1\%, 1.95\%, 4.23\% on the three applications respectively with 3.5x-6.4x memory reduction.
\end{abstract}




\maketitle

\section{Introduction}

Deep Neural Networks (DNNs) have been widely used in various applications and show its great potential to tackle complex problems \cite{blanco2019deep, chen2016deep, chen2016dcan, ronneberger2015u, yang2017suggestive, xu2018scaling}.
Furthermore, it has been and continue to be instrumental in enabling/advancing breakthroughs in various disciplines, including disease diagnosis, real-time language translation, autonomous driving, etc.\cite{xu2019whole, xu2020imagechd, zhang2017carcinopred, wang2020ica, gao2017quantitative,egger2017consumer,rosenberg2013improving,ji20123d, zhang2017carcinopred, sirinukunwattana2017gland}. 
%due to it has primarily been appraised according to classification / detection accuracy while traditional machine learning algorithms always have ceiling effect.
Meanwhile, edge computing for Internet of Things (IoT) has been widely studied which requires DNN models with small memory size and efficient computation \cite{xu2017edge, xu2018efficient,liu2019machine}. 
Thus, there is a huge gap between current DNN models and the requirenment of edge computing.
%, the deployment of DNN on edge requires smaller model sizes to fit limited device memory.% and maintain low latency to maintain user engagement.
%Talk about DNN performance and prevalance this paragraph. ALso mention their size issues.

Recently, in order to accommodate DNNs on the edge,  DNN quantization has become an active research topic \cite{ding2018universal, han2015deep, hubara2016quantized, li2016ternary,courbariaux2015binaryconnect, zhu2016trained, courbariauxbinarynet, hubara2016quantized, zhou2017incremental, rastegari2016xnor, zhou2016dorefa}, which aims to represent DNN weights with less memory (precision) while maintaining acceptable accuracy with efficient memory and computation costs.
It has been observed in the literature, however, that sometimes quantization can improve accuracy which can be credited to the reduction of overfitting \cite{srivastava2014dropout}.
%low-bit representation while maintaining an acceptable accuracy for efficient memory access and computation.
%Precision reduction.
Dynamic fixed point \cite{xu2018quantization} can achieve 4x less memory operation cost with only 0.4-0.6\% Top-5 accuracy loss for ImageNet classification \cite{deng2009imagenet}.
Ternary weight network and binaryConnect \cite{xu2018quantization} have further reduced the bit-width of weights to 2 bits or even 1 bit with a relatively larger accuracy loss.
Recently, their enhanced version, trained ternary training and binary weight network \cite{xu2018quantization} have reduced the accuracy loss to only 0.6-0.8\%.
There also exists some works using non-linear quantization to represent the parameter distribution for better accuracy \cite{xu2018quantization}.
Unlike the above works, some studies aims to quantize not only the weights but also the activations.
Quantized neural networks, binarized neural networks, and XNOR-net  \cite{xu2018quantization} reduced the weights to only 1 bit and the activations to 1-2 bits resulting in a large reduction on memory and computation cost yet with significant accuracy loss.
%Particularly, all the computation in XNOR-net are very simple XNOR operation.
Two-Step quantization \cite{wang2018two} has 2\% drop for VGG-16 on ImageNet-2012 dataset and Integer-Arithmetic-Only quantization\cite{jacob2018quantization} has 3.1\% accuracy loss.
In some of the above works, we notice that quantization can sometimes improve the performance \cite{xu2018quantization}, however, there is no comprehensive studies to verify this point. %however, there is no work .
%, which can be credited to the reduction of overfitting. 
%several works \cite{b6,wang2018two,jacob2018quantization,zhou2017incremental} discussed quantization, pruning, etc. to compress DNN model and simplify its computation.
%DNN quantization aims to represent its weights with fewer bits (low precision) while maintaining acceptable accuracy, efficient memory and computation cost.
%The main idea of it is to quantize the weights and/or activations of a DNN from 32-bit floating point into low-bit representations. 
%Discuss various quantization methods here. 
%
%It has been observed in the literature that sometimes quantization can improve accuracy which can be credited to the reduction of overfitting.

It is generally believed that quantization of DNN weights lead to performance degradation. For example, two-step quantization \cite{wang2018two} has a 2\% accuracy drop on ILSVRC-2012 dataset and Integer-Arithmetic-Only quantization \cite{jacob2018quantization} has a 3.1\% accuracy loss. Various works in the literature have explored the best ways to quantize the weights with minimum accuracy loss. 
To some extent, quantization essentially imposes regularization on weight representations. As such, it should also sometimes help to improve accuracy when overfitting issue presents in large neural networks.    

%\textcolor{ForestGreen}{

To demonstrate this point, in this paper, we conduct extensive experiments using incremental quantization on three applications: medical image segmentation, image classification and automatic speech recognition.
\textcolor{black}{For fair comparison, we have re-implemented related methods and performed evaluations under the same hardware, deep learning frameworks and configurations.}
In medical image segmentation, our method can achieve 1\% accuracy improvement compared with the current state-of-the-art method on the MICCIA Gland dataset.
In image classification, extensive experiments are presented on ImageNet-2012 using a widely used CNN model VGG-16 \cite{simonyan2014very}, and the result shows that our proposed method exceeds the current best performance using VGG-16 by up to 1.95\%.
In automatic speech recognition, we quantize Deep Speech \cite{hannun2014deep} network on the TIMIT dataset\cite{garofolo1993timit} and improves the accuracy by 4.23\%.
We also discuss the incremental quantization on the performance of simplified network with different bit widths, and the experimental results show that incremental quantization can no longer improve the performance of simplified networks with less overfitting.
In addition, we get 3.5x-6.4x memory reduction, which is extremely beneficial in the context of edge computing.

\begin{figure*}[!ht]
\begin{center}
%%\vspace{-10pt}
\centerline{\includegraphics[width=1\columnwidth]{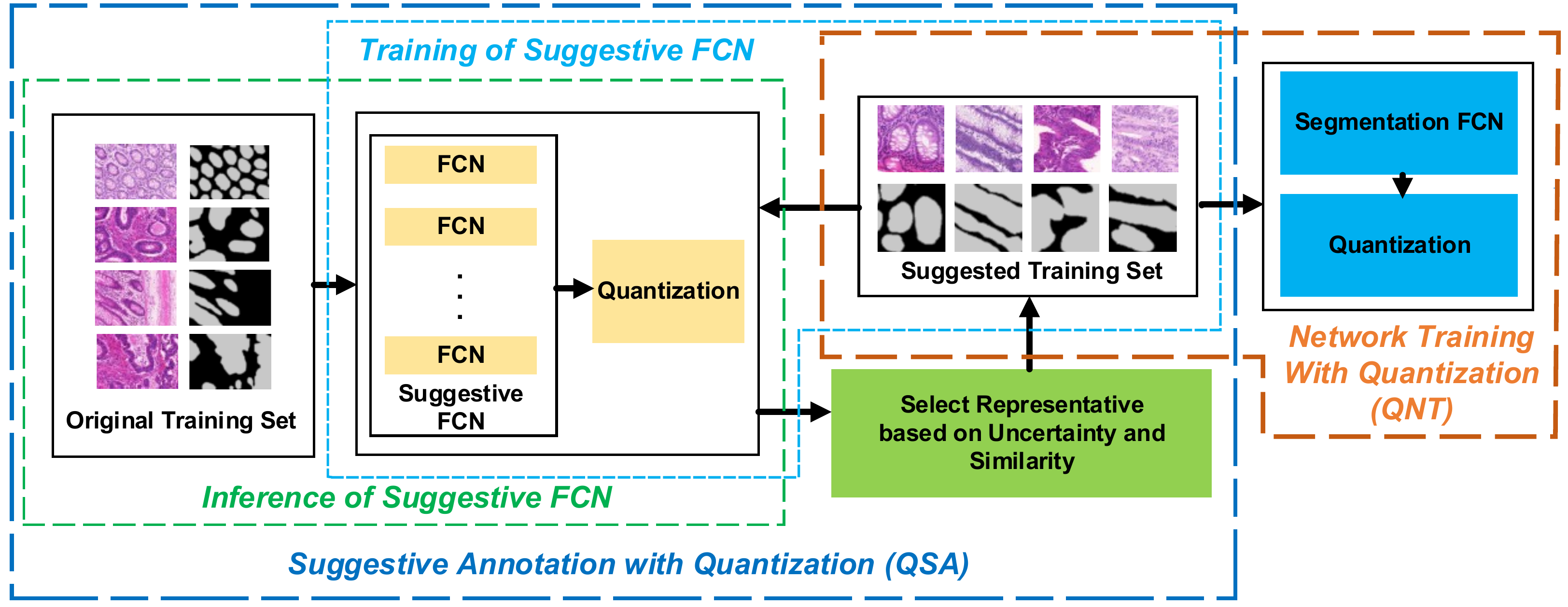}}
\end{center}
%\vspace{-30pt}
   \caption{Illustration of quantization framework based on the suggestive annotation framework. In suggestive annotation with quantization, better training samples (suggestive training set) can be extracted from the original training set. In network training with quantization, better performance can be achieved by reduce overfitting. }
\label{fig:architecture}
%%\vspace{-10pt}
\end{figure*}

\section{Preliminaries}
%please go through this section and merge the descriptions
In this section, we briefly review the incremental quantization method used in the experiments. 
The main goal of incremental quantization is to convert 32-bit-floating-point weights $W$ into low-precision weights $\hat{W}$ either power of two or zero with minimum accuracy loss.
Each of $\hat{W}$ is chosen from $P_{l}=\{\pm 2^{n_{1}}, \cdot \cdot \cdot, \pm 2^{n_{2}}\}$, where $n_{1}$ and $n_{2}$ are two integer numbers determined by the max absolute value of $W_{l}$ and expected quantized bit-width, and $n_{2}\leq n_{1}$.

The training procedure of incremental quantization \cite{zhou2017incremental} is consisted of three operations: weight partition, group-wise quantization and re-training.
Weights partition is the most crucial part of the whole process. 
%There are two widely used weight partition strategies: random partition and pruning-inspired partition.
We adopt pruning-inspired partition strategy to divide weights into two disjoint groups by comparing the absolute value with layer-wise threshold, which usually achieves better performance than random partition  (\cite{zhou2017incremental}).

For the $l$-th layer, we define weights partition as shown in Eq. (\ref{eq4.1.1}), where $A^{(1)}_{l}$ is the first group weights that need to be quantized and $A^{(2)}_{l}$ denotes the second group with remaining unquantized values.

\begin{equation}
\begin{aligned}
A^{(1)}_{l} \cup A^{(2)}_{l}&={W_{l}(i,j)}\\
A^{(1)}_{l} \cap A^{(2)}_{l}&=\emptyset
\end{aligned}
\label{eq4.1.1}
\end{equation}

The weights in the first group are expected to form a low-precision base for the original model, thus they are quantized using Eq. (\ref{eq2.1}),

\begin{equation}
w^{q}=\left\{
\begin{aligned}
&sign(w)\times 2^{p} \ \ \ \ if\ 3 \times 2^{p-2}  \leq  | w| \leq 3 \times 2^{p-1} ; \\
&sign(w)\times 2^{m}\ \ \ if\ |w| \geq 2^{u};\\
&0 \ \ \ \ \ \ \ \ \ \ \ \ \ \ \ \ \ \ \ if\ | w | < 2^{-l-1}
\end{aligned}
\right.
\label{eq2.1}
\end{equation}
where $w^{q}$ are quantized weights, $w$ represent original weights, $u$ and $l$ are upper and lower bounds of the quantized set($l \leq p \leq u$).

By contrast, the second group remains floating-point values to compensate the accuracy loss in model, and will be re-trained. 
After one round, these three steps are further adopted only on the second group in the rest of the training process.
As a result, all weights are quantized or to zeros hence we gain memory reduction with slight accuracy loss.

\section{Case Studies}
\subsection{Biomedical Image Segmentation}

\begin{table*}[!h]
\centering
%\vspace{-8pt}
\caption{Segmentation accuracy (averaged Dice score and F1 score in \%) using different quantization bit width on the MICCAI Gland dataset.}
%%\vspace{4pt}
\begin{tabular}{cccccccccc}
\hline
Configurations  & \multicolumn{9}{c}{Bit width}
%SA  Quantization bits
\\
& Orignal & 2 bits      & 3 bits     & 4 bits              & 5  bits             & 6 bits    & 7 bits      & 8 bits              & 9 bits              \\
\hline
Float SA + Quantized NT        & 86.25   & 65.55 &	79.37	& 85.87	& 86.12 &	86.33	& \textbf{86.51}	 & 85.93	& 85.26
          \\
%NT Quantization bits        & Orignal & 2 bits      & 3 bits     & 4 bits              & 5  bits             & 6 bits    & 7 bits      & 8 bits              & 9 bits              \\
Quantized SA + Float NT        &86.25 & 62.98 &	85.39	& 86.17	& 86.52	& \textbf{86.66}	& 86.51	& 86.22	& 86.12
   \\  \hline
\end{tabular}
\label{tab:seg_1}
%%\vspace{-8pt}
\end{table*}

\begin{table*}[!h]
\centering
%%\vspace{18pt}
\caption{Segmentation accuracy (averaged Dice score and F1 score in \%) using different parallel FCNs on the MICCAI Gland dataset.}
%%\vspace{4pt}
\begin{tabular}{cccccccc}
\hline
Configurations  & \multicolumn{7}{c}{Parallel number}
%SA  Quantization bits
\\
& Orignal &  2               & 3              & 4     & 5      & 6              & 7              \\\hline
Float SA + Float NT      & 86.25  & 86.23	& 85.93	& 85.81	& 86.24	& \textbf{87.25}	& 86.32    \\
Float SA + 7 bits NT &  86.25  & 86.2	& 85.39	& 85.32	& \textbf{86.84}	& 86.08	& 85.74    \\  \hline
\end{tabular}
\label{tab:seg_2}
%\vspace{-8pt}
\end{table*}

\begin{table*}[!h]
\centering
%%\vspace{18pt}
\caption{Segmentation accuracy (averaged Dice score and F1 score in \%) using different bit width and 5 parallel FCNs on the MICCAI Gland dataset.}
\begin{tabular}{cccccccccc}
\hline
Configurations  & \multicolumn{9}{c}{Bit width}
%SA  Quantization bits
\\
 & Orignal & 2 bits      & 3 bits     & 4 bits              & 5  bits             & 6 bits    & 7 bits      & 8 bits              & 9 bits     \\\hline
Float SA + 5 FCNs & 87.85 & 64.40	& 85.94	& 88.20	& 88.51	& 88.77	& \textbf{89.12} &	87.86	& 87.55 \\
7 bits SA + 5 FCNs      & 87.55   & 71.88 &	80.13	& 87.55	& 88.63	& 88.73	& \textbf{89.2}	& 88.67	& 87.53
         \\  \hline
\end{tabular}
\label{tab:seg_3}
%\vspace{-8pt}
\end{table*}

\subsubsection{Network Quantization}
As shown in Fig. \ref{fig:architecture}, the network quantization for biomedical image segmentation has two steps: suggestive annotation (SA) with quantization and network training (NT) with quantization.
In the first step, we add a quantization module to suggestive FCNs for high uncertainty.
In the second step, quantization of segmentation FCNs are performed with the suggestive training samples for higher accuracy.
The FCN is a 34 layer network, and more details can refer to \cite{yang2017suggestive}.
In order to obtain high representativeness, each FCN in suggestive FCNs should be diverse for high uncertainty with acceptable accuracy.
However, usually DNNs including FCNs are over-parameterized, and a large portion of the parameters is redundant.
Thus, multiple suggestive FCNs will have very small variance of the final prediction though with different weight initialization.
The adopted regularization techniques including weight decay and dropout scheme further make the multiple suggestive FCNs to be almost the same.
%In fact, we could use a relatively simple FCN as suggestive FCN.
%However, considering the minor difference on behaviours between different network structures and the extra cost, we tend to use the same FCNs for suggestive FCNs with segmentation FCNs.
%Popular regularization techniques should also be valid for robust models.
%Finally, we come up with the adoption of parameter quantization, which constraint the parameter space to defense overfitting problem.
By adding quantization to suggestive annotation, the above requirement can be satisfied.
Though it may be a little offensive since most of the time it will degrade the accuracy, it is particularly appreciated by suggestive FCNs that focus on uncertainty.
{\color{black}Particularly quantization transforms the originally continuous weight space, where the weights of several networks can be arbitrarily close, into a sparse and discrete one, and thus increasing the distances between the trained networks and accordingly the diversity of the outputs.}
Note that accuracy should be also considered and too offensive quantization methods should be avoided.
%\vspace{5pt}
\subsubsection{Experimental Setup}
We adopt the 2015 MICCAI Gland Challenge dataset \cite{sirinukunwattana2017gland} which has 85 training images (Part A: 37 normal glands, and Part B: 48 abnormal glands) and 80 testing images (Part A: 60 normal glands, and Part B: 20 abnormal glands).
In suggestive annotation \cite{yang2017suggestive}, 16 images with the highest uncertainty scores are extracted first, and then 8 images are collected based on their representativeness using similarity, which are added to the suggested training set in each iteration.
%{\color{black}Note that the training samples in the first iteration are selected randomly.}
Totally there are 120 iterations in suggestive annotation, and totally 960 suggested training samples are produced.
%\emph{Quantization bits:} 
We quantize the weights to 2 bits to 9 bits.
%, for these two configurations have the experimental best performance generally speaking.
%\emph{Ensembling numbers:} 
For ensembling method, We test 2 FCNs to 7 FCNs.
%to illustrate the affect of ensemble learning.
We also discuss the overfitting problem with a simplified version of FCN by reducing the filter numbers by half.
All the experiments are evaluated considering detection (F1 score) and segmentation (dice score) \cite{xu2018quantization}, and their averaged results are reported.

%\emph{Training parameters:} 
%5 FCNs are used in suggestive annotation,
The training parameters are as follows. 
We adopt a simple learning rate scaling strategy: set learning rate to 5$\times 10^{-4}$ in the initial stage, and to 5$\times 10^{-5}$ when the iteration time reaches a threshold.
All the configurations are repeated 4 times and the one with the optimal performance is selected for comparison.
%\vspace{5pt}
\subsubsection{Results and Analysis}
As there are two networks in suggestive annotation, we discussed both of the two with incremental quantization.
As shown in Table \ref{tab:seg_1}, we first analysis the performance of quantization methods in SA with quantization and NT with quantization. 
We can notice that compared with the original setting, quantization can improve the accuracy by 0.26\% and 0.31\% for quantized NT and quantized SA, respectively. 

As shown in Table \ref{tab:seg_2}, 
we can discover that ensembling method can improve the segmentation accuracy most of the time.
Similar to Table \ref{tab:seg_1}, the optimal accuracy is achieved in some median parallel number, e.g., 6 for float SA + 7 bits NT.

As shown in Table \ref{tab:seg_3}, we further discussed quantization on both SA and NT with the optimal parallel number.
The trend is the same as that in Table \ref{tab:seg_1}, and the median bit width (7 bits) obtains the best performance.
With both quantization and ensemble method, we can achieve a promising performance on the MICCAI 2015 Gland dataset, which outperforms the state-of-the-art method by 1\%. 
In addition, our method can also obtain 4.6x and 6.4x reduction on memory usage for incremental quantization with 7 bits and 5 bits, respectively.
Note that as activations are in floating point representation, the runtime are not affected.

As shown in Table \ref{tab:seg_small}, we also discuss quantization with simplified networks, e.g., small models.
We can notice that when the network model is smaller, the segmentation accuracy with 4 bits and 8 bits are no longer improved, however, there is a significant accuracy degradation.
This is due to the fact that smaller network with less parameters has less overfitting, and quantization further reduce the network's representation capability resulting with degraded performance.

%suggestive annotation (SA) with quantization has a great impact on the performance with one FCN, while network training with quantization has a significant influence on the performance with five FCNs.
%This is due to the fact that the network behaviour of multiple networks with ensemble methods differs from that of only one network.

%Then we show the performance comparison of quantization methods in network training with quantization with training samples from suggestive annotation and suggestive annotation with quantization, which gets 6 bits quantization network training.
%Third part, we compare different number of suggestive FCNs with network training and network training with quantization, and get 1\% and 0.59\% improvement respectively.
%Last part, we compare of quantization methods in network training with quantization using training samples from suggestive annotation (SA) and suggestive annotation with quantization as well as ensemble learning.
%All in all, our methods can achieve a 0.9\%-1\% improvement with the original method.
\begin{table}[!h]
\centering
\caption{Segmentation accuracy (averaged Dice score and F1 score in \%) using quantization with 4 bits and 8 bits with small models.}
\begin{tabular}{cccc}
\hline
 Network        & float & 4 bit & 8 bit \\ \hline
simplified (Quantized SA + Float NT)             & \textbf{82.53} & {78.14} & 82.14 \\
simplified (Float SA + Quantized NT)       & \textbf{82.53} & 76.12 & {81.27} \\  \hline
\end{tabular}
\label{tab:seg_small}
\end{table}

%\vspace{-5pt}
\subsection{Image Classification}

\subsubsection{Network Quantization}

\begin{figure*}[ht]
\begin{center}
%\vspace{-10pt}
\centerline{\includegraphics[width=1\columnwidth]{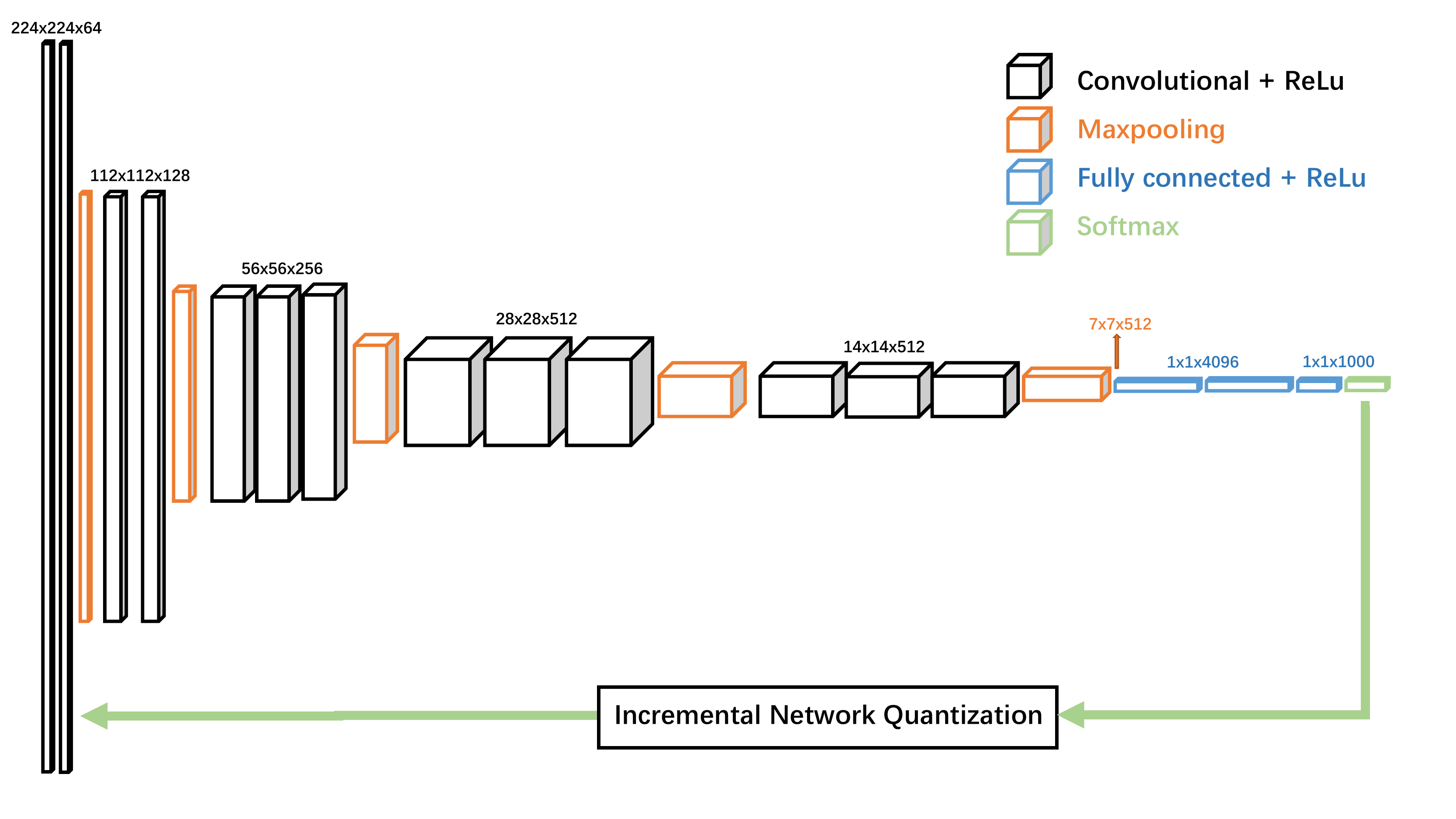}}
\end{center}
%\vspace{-20pt}
   \caption{Illustration of quantization framework on the VGG-16 model. In small VGG models, we remove the first layers (with a feature map size of 112x112x128) and halved the channels of the rest layers except for the first two input layers. In small VGG-16 models with 5 layers, we remove the third and forth layers (with  feature map sizes of 28x28x512 and 14x14x512, respectively). In small VGG-16 models with eight layers, we only remove the forth layer. }
\label{fig:archi_vgg}
%\vspace{-1pt}
\end{figure*}

We use incremental quantization \cite{zhou2017incremental} on VGG-16 model for image classification. 
The original VGG-16 has 16 layers (13 conv-layers and 3 FC layers) as shown in Fig. \ref{fig:archi_vgg}.
Besides that, we focus on the parameter-redundancy problem on VGG-16.
We re-train two small VGG-16 models with the first five convolutional layers of the original eight layers) and the first eight convolutional layers of the original 11 layers), respectively for overfitting analysis.
%And the results are shown as blow.
%%\vspace{5pt}
\subsubsection{Experimental Setup}
We adopt the ILSVRC-2012 dataset \cite{deng2009imagenet} which contains over 1.2 million training images and over 50,000 testing images. VGG-16 \cite{simonyan2014very} is adopted for classification.
We quantize the weights from 2 bits to 7 bits for discussion.
Additionally, we test two VGG models to 9 VGG models to further clarify the compensation of ensemble learning in terms of accuracy loss brought by weight quantization.
And for small VGG-16 models, we compare 4 bits quantization and 8 bits quantization with floating weights.
Learning rate is set to $1\times 10^{-4}$ with learning rate decay as $1\times 10^{-6}$.
Stochastic gradient descent (SGD) function is selected to be the optimizer and Momentun of 0.9 is used to accelerate training.
In order to unify the variables and discuss how the bit width impact the network performance, the max value of the quantized weights is set to 4.0, and it takes 4 iterations from full floating-point weights to quantized weights, and the accumulated portions of quantized weights at four iterative steps are \{50\%, 75\%, 87.5\%, 100\%\}.
%%\vspace{5pt}

\begin{table*}[ht]
\centering
%\vspace{-1pt}
\caption{Image classification performance (top-1 error in \%) of VGG-16 model \cite{jacob2018quantization} with weights from 2 bits to 9 bits on ImageNet dataset.}
\begin{tabular}{cccccccccc}
\hline
Configurations  & \multicolumn{9}{c}{Quantization bit width}\\
     & Original & 2 bits      & 3 bits     & 4 bits              & 5  bits             & 6 bits    & 7 bits      & 8 bits              & 9 bits              \\ \hline
   VGG-16      & 69.29   & 42.61  & 63.72  & 67.70            & 68.64          & 69.68 & 70.36  & \textbf{70.48} & 70.28          \\ \hline
\end{tabular}
\label{tab:vgg16}
\end{table*}

\begin{table*}[!htbp]
\centering
%\vspace{-10pt}
\caption{Image classification performance (top-1 error in \%) of VGG-16 model \cite{jacob2018quantization} with parallel numbers from 2 to 7 on ImageNet dataset.}
\begin{tabular}{ccccccccc}
\hline
Configurations  & \multicolumn{8}{c}{parallel numbers}\\
       & Orignal &  2      & 3               & 4              & 5     & 6      & 7        \\ \hline
       VGG-16  & 69.29    & 70.18  & \textbf{70.10} & 70.38         & 69.31 & 69.82 & 68.58          &                \\
       VGG-16 with 8 bits      & 69.29   & 70.10 & \textbf{71.24}  & 71.03         & 69.80  & 69.17  & 69.85         \\      \hline
\end{tabular}
\label{tab:vgg16_ens}
\end{table*}

\begin{table}[ht]
\centering
\caption{Classification performance (top-1 error in \%) comparison between floating pointss, 4 bits and 8 bits representations
using small VGG-16 models on ImageNet.}
\begin{tabular}{cccc}
\hline
 Network        & float & 4 bit & 8 bit \\ \hline
VGG-16 with 5 conv-layers             & \textbf{64.14} & 64.12 & 64.05 \\
VGG-16 with 8 conv-layers           & \textbf{66.04} & 66.00 & 66.02 \\\hline
\end{tabular}
\label{tab:vgg16_small}
\end{table}

% \begin{table*}[!ht]
% \centering
% \caption{Image classification performance on 8 bits VGG-16 model \cite{jacob2018quantization} with parallel numbers from 2 to 7.}
% \begin{tabular}{cccccccccc}
% \hline
% Configurations  & \multicolumn{9}{c}{parallel numbers}\\
%         & Orignal & Best Q & Best E & 2               & 3              & 4     & 5      & 6              & 7              \\ \hline
% VGG-16 with 8 bits      & 69.29   & 70.48  & 70.10 & \textbf{71.24}  & 71.03         & 69.80  & 69.17  & 69.85          & 69.23          \\  \hline
% \end{tabular}
% \label{tab:vgg16_ens_q}
% \end{table*}

\subsubsection{Results and Analysis}
We compare our result with the original and small VGG-16 models.
As shown in Table \ref{tab:vgg16}, 
%we first analysis the performance of quantization on original VGG-16 model from 2 bits to 9 bits.
%the result of quantization experiment from 2 bits to 9 bits is listed in Row 2.
the configuration with 8 bits obtains the optimal performance, which gets a 1.1\% improvement.
We set a variety of ensemble experiments as the control group to discuss the effect of ensemble methods.
The result of six configurations from two to seven parallel models using 32-bit floating-point weights are shown in Table \ref{tab:vgg16_ens}.
With three parallel models, we can get a 0.81\% performance improvement.
%In Table. \ref{tab:vgg16_ens_q}, 
%We finally utilize 
While for the VGG-16 model with  8 bits representation and parallel number from two to seven, we get a 1.95\% improvement with 2 parallel models compared with the original model.
In addition, 4x memory reduction is obtained with quantization.

As for overfitting, we discuss the performance of 4 bits and 8 bits quantization on small VGG-16 models.
We shorten the original VGG-16 model to contain 5/8 conv-layers with 3 FC layers.
As shown in Table. \ref{tab:vgg16_small}, we get 0.02\% precision loss after quantization on both two simplified VGG-16 models.
Comparing to the results in Table. \ref{tab:vgg16}, it can be concluded that quantization can benefit the over-parameterized models to some extent.
However, if the original model is already simplified, the inhibiting effect of quantization will be faded.

\begin{figure*}[!h]
\begin{center}
%\vspace{-10pt}
\centerline{\includegraphics[width=1\columnwidth]{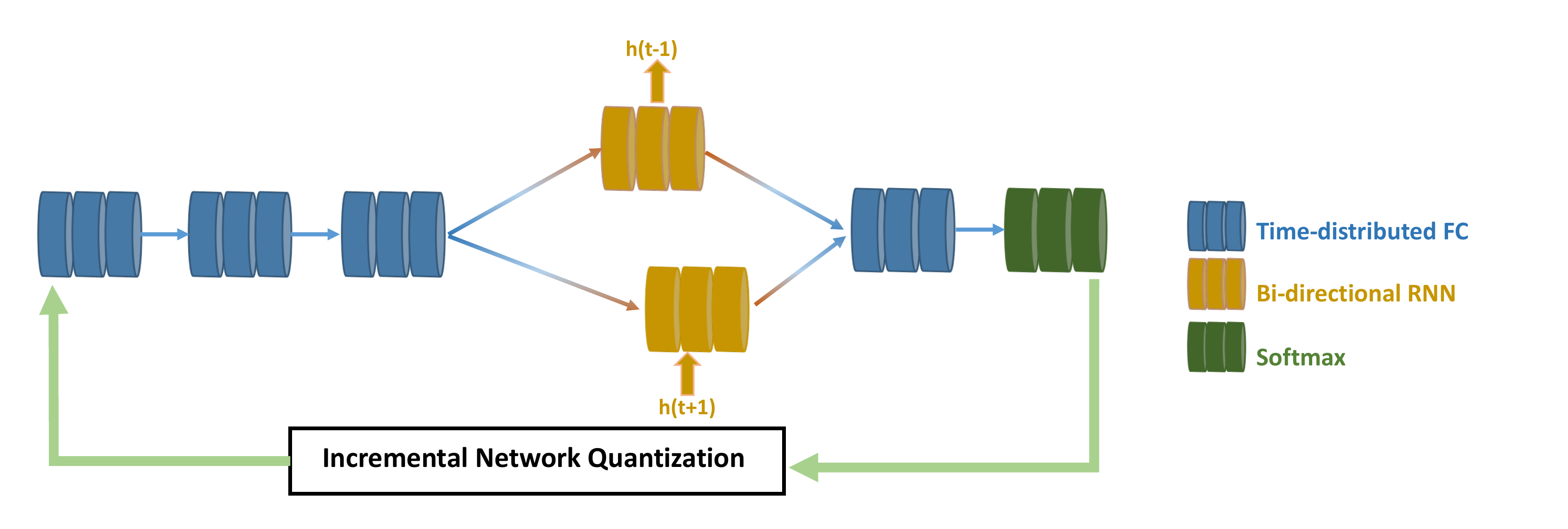}}
\end{center}
%\vspace{-20pt}
   \caption{Illustration of quantization framework on the Deep Speech. For small Deep Speech models, we remove the second and third FC layers. }
\label{archi_ds}
%%\vspace{-10pt}
\end{figure*}

\begin{table*}[!h]
\centering
\caption{Automatic speech  recognition performance (classification error rate in \%) on Deep Speech \cite{zhou2017incremental} with weights from 2 bits to 9 bits.}
\begin{tabular}{cccccccccc}
\hline
Configurations  & \multicolumn{9}{c}{Quantization bit width} \\
                 & Original & 2 bits      & 3 bits     & 4 bits              & 5  bits             & 6 bits    & 7 bits      & 8 bits              & 9 bits              \\\hline
Deep Speech    & 48.13   & 60.33  & 64.30  & 69.53           & 52.93          & 47.20 & 49.50  & 52.40          & \textbf{44.70} \\ \hline
\end{tabular}
\label{tab:ds}
\end{table*}

\begin{table*}[!h]
\centering
\caption{Automatic speech recognition performance (classification error rate in \%) on Deep Speech \cite{zhou2017incremental} with parallel numbers from 2 to 7.}
\begin{tabular}{ccccccccc}
\hline
Configurations  & \multicolumn{8}{c}{Parallel numbers}\\
        & Orignal &  2      & 3               & 4              & 5     & 6      & 7              &                \\ \hline
Deep Speech   & 48.13  & 45.37  & 47.30           & \textbf{43.93} & 42.83 & 46.23  & 43.30          &                \\ 
Deep Speech with 9 bits & 48.13  & \textbf{43.90}  & 46.17          & 48.07 & 48.83  & 53.90          & 45.53 \\
\hline
\end{tabular}
\label{tab:ds_ens}
\end{table*}

\begin{table}[!h]
\centering
\caption{Automatic speech recognition performance (classification error rate in \%) comparison between floating pointss, 4 bits and 8 bits representations
using small Deep Speech models}
\begin{tabular}{cccc}
\hline
 Network        & float & 4 bit & 8 bit \\ \hline
Small Deep Speech & 50.04 & \textbf{47.05} & 47.08 \\ \hline
\end{tabular}
\label{tab:ds_small}
\end{table}

% \begin{table*}[!ht]
% \centering
% \caption{Automatic speech  recognition performance on 9 bits Deep Speech \cite{zhou2017incremental} with parallel numbers from 2 to 7.}
% \begin{tabular}{cccccccccc}
% \hline
% Configurations  & \multicolumn{9}{c}{parallel numbers}\\
%         & Orignal & Best Q & Best E & 2               & 3              & 4     & 5      & 6              & 7              \\ \hline
% Deep Speech with 9 bits & 48.13   & 44.70  & 43.93  & \textbf{43.90}  & 46.17          & 48.07 & 48.83  & 53.90          & 45.53          \\ \hline
% \end{tabular}
% \label{tab:ds_ens_q}
% \end{table*}

%%\vspace{5pt}
\subsection{Automatic Speech Recognition}

\subsubsection{Network Quantization}

We use incremental quantization \cite{zhou2017incremental} on Deep Speech for speech recognition task. 
As shown in Fig. \ref{archi_ds}, the original Deep Speech has 5 hidden layers, composed of 3 time-distributed FC layers, 1 bi-directional RNN layer, 1 time-distributed FC layer followed by 1 softmax layer as output.
%The over-paramaterize problem also disccussed in this task.
We remove 2 time-distributed FC layers before the RNN layer to simplify the model, and re-train the small model on TIMIT.
%%\vspace{5pt}
\subsubsection{Experiment Setup}
We adopt the DARPA TIMIT Acoustic-Phonetic Continuous Speech Corpus dataset (TIMIT) \cite{garofolo1993timit} which contains 6300 sentences read by 630 different Americans for Deep Speech (DS) network \cite{hannun2014deep}.
%This experiment makes a further verification for the generalizable of our method.
We quantize the weights to 2 bits to 7 bits, and we test two to nine models in ensemble method.
For training parameters, we set learning rate to $1\times 10^{-4}$ with a learning rate decay of $1\times 10^{-6}$.
%Also, nesterov accelerated gradient is used to be the optimizer.
The predicted error rate will be evaluated by the word error rate (WER).

%%\vspace{5pt}
\subsubsection{Results and Analysis}
%For the speech recognition task, 
As shown in Table \ref{tab:ds}, the quantized model with 9 bits obtains the optimal performance which is 3.43\% higher than the model with floating point representation.
As shown in Table. \ref{tab:ds_ens}, With three parallel networks, we can get an improvement of 4.2\%. Using two parallel models, our method obtains the optimal performance among all the configurations which is 4.23\% higher than the original method, which is 0.03\% better than the ensemble model.
In addition, 3.6x reduction on memory usage can be achieved.

%The whole experiment reveals that our method is robust and be able to generalize to other network structure.
In Table. \ref{tab:ds_small}, we discuss the quantization effectiveness on small Deep Speech models using 4 bits and 8 bits.
Our model can get an improvement of 3\% compared with the original model.

%statement of conclusion
%\vspace{5pt}
\subsection{Discussion}

As for the accuracy, we can notice that incremental quantization can improve the segmentation performance by around 1\% on the MICCIA Gland dataset. For image classification, We ended up using the 8-bit fixed-point VGG-16 model for two parallel models to achieve a 1.95\%
improvement compared with the original model. For speech recognition, the optimal performance obtained an improvement of 4.23\% over the original Deep Speech model. In addition, our method has a up to 3.5-6.4x reduction on memory usage.
%In this part, we will discuss some abnormal situations and give the summary of the whole experiment.
By comparing network training with and without quantization in Table \ref{tab:seg_1}, Table \ref{tab:vgg16} and Table \ref{tab:ds}, we can observe that network training with quantization will not always improve the accuracy, especially in extremely situations with 2-4 bits representation.
In addition, too large bit widths do not improve the performance, and the optimal performance is achieved with some median bit width, e.g., 6 bits and 7 bits.
Thus, with proper incremental quantization, the accuracy can usually be improved with memory reduction.
%The trends sho w that we can get better performance by using larger bitwidth.
%However, considering the cost on memory as one of our main concerns, we only set the widest bitwidth to 9.

As for the impact on ensemble method, six parallel FCNs, two parallel CNNs, and two parallel RNNs with quantization reaches the optimal performance on segmentation, classification and automatic speech recognition, respectively.
Also the optimal accuracy is achieved in some median parallel number.
Thus, a proper parallel number can improve the accuracy by scarifying some memory operation and computation cost.
%quantization method can boost the performance of the whole framework in terms of accuracy and memory reduction.

\textcolor{black}{For overfitting, we discuss the corresponding small models in the three applications with quantiztaion. As shown in Table \ref{tab:seg_small}, Table \ref{tab:vgg16_small} and Table \ref{tab:ds_small}, we can notice that when the model is much smaller than the original networks, the accuracy degrades seriously such as the results in medical image segmentation and image classification.
When the model is only simplified a bit such as the small model in automatic speech recognition, the accuracy is still improved.
This is possibly due to the fact that large models usually have more overfitting than small ones.
Thus, when quantization is applied, the overfitting in large models is reduced, resulting with improved performance.
However, for small models with less overfitting, the overfitting is reduced and the representation capability is also degraded resulting with accuracy loss.
Therefore, incremental quantization may be used as a regulation method to reduce overfitting.}

\textcolor{black}{
Incremental quantization can not only reduce memory consumption of 3.5x-6.4x, but also speedup the processing. 
As the optimization mainly comes from transforming multiplication to addition in the convolution calculation, specific hardware modules are required for speedup.
Thus, the speedup depends on specific hardware such as CPU, GPU and FPGAs.
If no specific hardware module is implemented for such transforming, no speedup is obtained, e.g., on existing CPUs and GPUs. 
If we implement specific hardware module for 2D convolution operation on FPGAs, the speedup can be 1.7x-7.8x according to \cite{xu2017edge}. Note that the speedup depends on a variety of factors such as hardware platforms and network structures.
}
%to some extent quantization is helpful to reduce the redundancy in large scale models. 
%Results shows that for small FCN networks, quantization hurts the performance rather than restrict the overfitting.
%And for the original small networks such as VGG-16 and Deep Speech, quantization can still benefit the network regardless of the scale.

%  This experiment has proved that
% our framework is robust and can be adopt to multiple kinds of
% tasks, and less limited by data features and network structures.
% Our future work will focus on a more effective quantization
% method to obtain better performance after quantization, which
% will rely less on ensemble technique and have smaller volume
% for hardware applications.
 
\section{Conclusion}
Among deep neural networks (DNNs), compression techniques such as weight quantization and pruning are usually applied before they can be  accommodated on the edge. It is generally believed that quantization leads to performance degradation, and plenty of existing works have explored quantization strategies aiming at minimum accuracy loss. In this paper, we show that quantization can sometimes help to improve accuracy by imposing regularization on weight representations. 
We conduct comprehensive experiments on three widely used applications: fully connected network (FCN) for biomedical image segmentation, convolutional neural network (CNN) for image classification, and recurrent neural network (RNN) for automatic speech recognition, and
experimental results show that incremental quantization can improve the accuracy by 1\%, 1.95\%, 4.23\% on the three applications respectively with 3.5x-6.4x memory reduction.
\textcolor{black}{As a case in compression techniques, incremental quantization shows great potential to reduce onverfitting, and there may exist some general rules and strategies to enable other compression techniques to have the capability of reducing overfitting. 
We encourage related researchers to explore this interesting topic, which is also our future work.
}

\begin{acks}
This work was supported by the National key Research and Development Program of China (No. 2018YFC1002600), the Science and Technology Planning Project of Guangdong Province, China (No. 2017B090904034, No. 2017B030314109, No. 2018B090944002, No. 2019B020230003), Guangdong Peak Project (No. DFJH201802), the National Natural Science Foundation of China (No. 62006050).

\end{acks}

\bibliographystyle{ACM-Reference-Format}
\bibliography{sample-base}


\begin{thebibliography}{36}

%%% the Web address.

\ifx \showCODEN    \undefined \def \showCODEN     #1{\unskip}     \fi
\ifx \showDOI      \undefined \def \showDOI       #1{#1}\fi
\ifx \showISBNx    \undefined \def \showISBNx     #1{\unskip}     \fi
\ifx \showISBNxiii \undefined \def \showISBNxiii  #1{\unskip}     \fi
\ifx \showISSN     \undefined \def \showISSN      #1{\unskip}     \fi
\ifx \showLCCN     \undefined \def \showLCCN      #1{\unskip}     \fi
\ifx \shownote     \undefined \def \shownote      #1{#1}          \fi
\ifx \showarticletitle \undefined \def \showarticletitle #1{#1}   \fi
\ifx \showURL      \undefined \def \showURL       {\relax}        \fi
% invisible to TeX
\providecommand\bibfield[2]{#2}
\providecommand\bibinfo[2]{#2}
\providecommand\natexlab[1]{#1}
\providecommand\showeprint[2][]{arXiv:#2}

\bibitem[\protect\citeauthoryear{Blanco-Filgueira, Garc{\'\i}a-Lesta,
  Fern{\'a}ndez-Sanjurjo, Brea, and L{\'o}pez}{Blanco-Filgueira
  et~al\mbox{.}}{2019}]%
        {blanco2019deep}
\bibfield{author}{\bibinfo{person}{Beatriz Blanco-Filgueira},
  \bibinfo{person}{Daniel Garc{\'\i}a-Lesta}, \bibinfo{person}{Mauro
  Fern{\'a}ndez-Sanjurjo}, \bibinfo{person}{V{\'\i}ctor~Manuel Brea}, {and}
  \bibinfo{person}{Paula L{\'o}pez}.} \bibinfo{year}{2019}\natexlab{}.
\newblock \showarticletitle{Deep learning-based multiple object visual tracking
  on embedded system for iot and mobile edge computing applications}.
\newblock \bibinfo{journal}{\emph{IEEE Internet of Things Journal}}
  \bibinfo{volume}{6}, \bibinfo{number}{3} (\bibinfo{year}{2019}),
  \bibinfo{pages}{5423--5431}.
\newblock


\bibitem[\protect\citeauthoryear{Chen, Qi, Cheng, Heng, et~al\mbox{.}}{Chen
  et~al\mbox{.}}{2016a}]%
        {chen2016deep}
\bibfield{author}{\bibinfo{person}{Hao Chen}, \bibinfo{person}{Xiaojuan Qi},
  \bibinfo{person}{Jie-Zhi Cheng}, \bibinfo{person}{Pheng-Ann Heng},
  {et~al\mbox{.}}} \bibinfo{year}{2016}\natexlab{a}.
\newblock \showarticletitle{Deep Contextual Networks for Neuronal Structure
  Segmentation.}. In \bibinfo{booktitle}{\emph{AAAI}}.
  \bibinfo{pages}{1167--1173}.
\newblock


\bibitem[\protect\citeauthoryear{Chen, Qi, Yu, and Heng}{Chen
  et~al\mbox{.}}{2016b}]%
        {chen2016dcan}
\bibfield{author}{\bibinfo{person}{Hao Chen}, \bibinfo{person}{Xiaojuan Qi},
  \bibinfo{person}{Lequan Yu}, {and} \bibinfo{person}{Pheng-Ann Heng}.}
  \bibinfo{year}{2016}\natexlab{b}.
\newblock \showarticletitle{Dcan: Deep contour-aware networks for accurate
  gland segmentation}. In \bibinfo{booktitle}{\emph{CVPR}}.
  \bibinfo{pages}{2487--2496}.
\newblock


\bibitem[\protect\citeauthoryear{Courbariaux and Bengio}{Courbariaux and
  Bengio}{[n.d.]}]%
        {courbariauxbinarynet}
\bibfield{author}{\bibinfo{person}{M Courbariaux} {and} \bibinfo{person}{Y
  Bengio}.} \bibinfo{year}{[n.d.]}\natexlab{}.
\newblock \bibinfo{title}{Binarynet: Training deep neural networks with weights
  and activations constrained to+ 1 or-1. CoRR abs/1602.02830 (2016)}.
\newblock
\newblock


\bibitem[\protect\citeauthoryear{Courbariaux, Bengio, and David}{Courbariaux
  et~al\mbox{.}}{2015}]%
        {courbariaux2015binaryconnect}
\bibfield{author}{\bibinfo{person}{Matthieu Courbariaux},
  \bibinfo{person}{Yoshua Bengio}, {and} \bibinfo{person}{Jean-Pierre David}.}
  \bibinfo{year}{2015}\natexlab{}.
\newblock \showarticletitle{Binaryconnect: Training deep neural networks with
  binary weights during propagations}. In \bibinfo{booktitle}{\emph{NIPS}}.
  \bibinfo{pages}{3123--3131}.
\newblock


\bibitem[\protect\citeauthoryear{Deng, Dong, Socher, Li, Li, and Fei-Fei}{Deng
  et~al\mbox{.}}{2009}]%
        {deng2009imagenet}
\bibfield{author}{\bibinfo{person}{Jia Deng}, \bibinfo{person}{Wei Dong},
  \bibinfo{person}{Richard Socher}, \bibinfo{person}{Li-Jia Li},
  \bibinfo{person}{Kai Li}, {and} \bibinfo{person}{Li Fei-Fei}.}
  \bibinfo{year}{2009}\natexlab{}.
\newblock \showarticletitle{Imagenet: A large-scale hierarchical image
  database}. In \bibinfo{booktitle}{\emph{CVPR}}. IEEE,
  \bibinfo{pages}{248--255}.
\newblock


\bibitem[\protect\citeauthoryear{Ding, Liu, and Shi}{Ding
  et~al\mbox{.}}{2018}]%
        {ding2018universal}
\bibfield{author}{\bibinfo{person}{Yukun Ding}, \bibinfo{person}{Jinglan Liu},
  {and} \bibinfo{person}{Yiyu Shi}.} \bibinfo{year}{2018}\natexlab{}.
\newblock \showarticletitle{On the Universal Approximability of Quantized ReLU
  Neural Networks}.
\newblock \bibinfo{journal}{\emph{arXiv preprint arXiv:1802.03646}}
  (\bibinfo{year}{2018}).
\newblock


\bibitem[\protect\citeauthoryear{Egger and Schoder}{Egger and Schoder}{2017}]%
        {egger2017consumer}
\bibfield{author}{\bibinfo{person}{Marc Egger} {and} \bibinfo{person}{Detlef
  Schoder}.} \bibinfo{year}{2017}\natexlab{}.
\newblock \showarticletitle{Consumer-oriented tech mining: Integrating the
  consumer perspective into organizational technology intelligence-the case of
  autonomous driving}. In \bibinfo{booktitle}{\emph{Proceedings of the 50th
  Hawaii International Conference on System Sciences}}.
\newblock


\bibitem[\protect\citeauthoryear{Gao, Wang, Zhang, Hu, Yang, Wang, Zhang, and
  Wang}{Gao et~al\mbox{.}}{2017}]%
        {gao2017quantitative}
\bibfield{author}{\bibinfo{person}{Ge Gao}, \bibinfo{person}{Chengyan Wang},
  \bibinfo{person}{Xiaodong Zhang}, \bibinfo{person}{Juan Hu},
  \bibinfo{person}{Xuedong Yang}, \bibinfo{person}{He Wang},
  \bibinfo{person}{Jue Zhang}, {and} \bibinfo{person}{Xiaoying Wang}.}
  \bibinfo{year}{2017}\natexlab{}.
\newblock \showarticletitle{Quantitative analysis of diffusion-weighted
  magnetic resonance images: differentiation between prostate cancer and normal
  tissue based on a computer-aided diagnosis system}.
\newblock \bibinfo{journal}{\emph{Science China Life Sciences}}
  \bibinfo{volume}{60}, \bibinfo{number}{1} (\bibinfo{year}{2017}),
  \bibinfo{pages}{37--43}.
\newblock


\bibitem[\protect\citeauthoryear{Garofolo}{Garofolo}{1993}]%
        {garofolo1993timit}
\bibfield{author}{\bibinfo{person}{John~S Garofolo}.}
  \bibinfo{year}{1993}\natexlab{}.
\newblock \showarticletitle{TIMIT acoustic phonetic continuous speech corpus}.
\newblock \bibinfo{journal}{\emph{Linguistic Data Consortium, 1993}}
  (\bibinfo{year}{1993}).
\newblock


\bibitem[\protect\citeauthoryear{Han, Mao, and Dally}{Han
  et~al\mbox{.}}{2015}]%
        {han2015deep}
\bibfield{author}{\bibinfo{person}{Song Han}, \bibinfo{person}{Huizi Mao},
  {and} \bibinfo{person}{William~J Dally}.} \bibinfo{year}{2015}\natexlab{}.
\newblock \showarticletitle{Deep compression: Compressing deep neural networks
  with pruning, trained quantization and huffman coding}.
\newblock \bibinfo{journal}{\emph{arXiv preprint arXiv:1510.00149}}
  (\bibinfo{year}{2015}).
\newblock


\bibitem[\protect\citeauthoryear{Hannun, Case, Casper, Catanzaro, Diamos,
  Elsen, Prenger, Satheesh, Sengupta, Coates, et~al\mbox{.}}{Hannun
  et~al\mbox{.}}{2014}]%
        {hannun2014deep}
\bibfield{author}{\bibinfo{person}{Awni Hannun}, \bibinfo{person}{Carl Case},
  \bibinfo{person}{Jared Casper}, \bibinfo{person}{Bryan Catanzaro},
  \bibinfo{person}{Greg Diamos}, \bibinfo{person}{Erich Elsen},
  \bibinfo{person}{Ryan Prenger}, \bibinfo{person}{Sanjeev Satheesh},
  \bibinfo{person}{Shubho Sengupta}, \bibinfo{person}{Adam Coates},
  {et~al\mbox{.}}} \bibinfo{year}{2014}\natexlab{}.
\newblock \showarticletitle{Deep speech: Scaling up end-to-end speech
  recognition}.
\newblock \bibinfo{journal}{\emph{arXiv preprint arXiv:1412.5567}}
  (\bibinfo{year}{2014}).
\newblock


\bibitem[\protect\citeauthoryear{Hubara, Courbariaux, Soudry, El-Yaniv, and
  Bengio}{Hubara et~al\mbox{.}}{2016}]%
        {hubara2016quantized}
\bibfield{author}{\bibinfo{person}{Itay Hubara}, \bibinfo{person}{Matthieu
  Courbariaux}, \bibinfo{person}{Daniel Soudry}, \bibinfo{person}{Ran
  El-Yaniv}, {and} \bibinfo{person}{Yoshua Bengio}.}
  \bibinfo{year}{2016}\natexlab{}.
\newblock \showarticletitle{Quantized neural networks: Training neural networks
  with low precision weights and activations}.
\newblock \bibinfo{journal}{\emph{arXiv preprint arXiv:1609.07061}}
  (\bibinfo{year}{2016}).
\newblock


\bibitem[\protect\citeauthoryear{Jacob, Kligys, Chen, Zhu, Tang, Howard, Adam,
  and Kalenichenko}{Jacob et~al\mbox{.}}{2018}]%
        {jacob2018quantization}
\bibfield{author}{\bibinfo{person}{Benoit Jacob}, \bibinfo{person}{Skirmantas
  Kligys}, \bibinfo{person}{Bo Chen}, \bibinfo{person}{Menglong Zhu},
  \bibinfo{person}{Matthew Tang}, \bibinfo{person}{Andrew Howard},
  \bibinfo{person}{Hartwig Adam}, {and} \bibinfo{person}{Dmitry Kalenichenko}.}
  \bibinfo{year}{2018}\natexlab{}.
\newblock \showarticletitle{Quantization and training of neural networks for
  efficient integer-arithmetic-only inference}. In
  \bibinfo{booktitle}{\emph{Proceedings of the IEEE Conference on Computer
  Vision and Pattern Recognition}}. \bibinfo{pages}{2704--2713}.
\newblock


\bibitem[\protect\citeauthoryear{Ji, Xu, Yang, and Yu}{Ji
  et~al\mbox{.}}{2012}]%
        {ji20123d}
\bibfield{author}{\bibinfo{person}{Shuiwang Ji}, \bibinfo{person}{Wei Xu},
  \bibinfo{person}{Ming Yang}, {and} \bibinfo{person}{Kai Yu}.}
  \bibinfo{year}{2012}\natexlab{}.
\newblock \showarticletitle{3D convolutional neural networks for human action
  recognition}.
\newblock \bibinfo{journal}{\emph{IEEE transactions on pattern analysis and
  machine intelligence}} \bibinfo{volume}{35}, \bibinfo{number}{1}
  (\bibinfo{year}{2012}), \bibinfo{pages}{221--231}.
\newblock


\bibitem[\protect\citeauthoryear{Li, Zhang, and Liu}{Li et~al\mbox{.}}{2016}]%
        {li2016ternary}
\bibfield{author}{\bibinfo{person}{Fengfu Li}, \bibinfo{person}{Bo Zhang},
  {and} \bibinfo{person}{Bin Liu}.} \bibinfo{year}{2016}\natexlab{}.
\newblock \showarticletitle{Ternary weight networks}.
\newblock \bibinfo{journal}{\emph{arXiv preprint arXiv:1605.04711}}
  (\bibinfo{year}{2016}).
\newblock


\bibitem[\protect\citeauthoryear{Liu, Xu, Liu, Liu, Wang, Shi, Wen, Huang,
  Yuan, and Zhuang}{Liu et~al\mbox{.}}{2019}]%
        {liu2019machine}
\bibfield{author}{\bibinfo{person}{Zihao Liu}, \bibinfo{person}{Xiaowei Xu},
  \bibinfo{person}{Tao Liu}, \bibinfo{person}{Qi Liu}, \bibinfo{person}{Yanzhi
  Wang}, \bibinfo{person}{Yiyu Shi}, \bibinfo{person}{Wujie Wen},
  \bibinfo{person}{Meiping Huang}, \bibinfo{person}{Haiyun Yuan}, {and}
  \bibinfo{person}{Jian Zhuang}.} \bibinfo{year}{2019}\natexlab{}.
\newblock \showarticletitle{Machine vision guided 3d medical image compression
  for efficient transmission and accurate segmentation in the clouds}. In
  \bibinfo{booktitle}{\emph{Proceedings of the IEEE/CVF Conference on Computer
  Vision and Pattern Recognition}}. \bibinfo{pages}{12687--12696}.
\newblock


\bibitem[\protect\citeauthoryear{Rastegari, Ordonez, Redmon, and
  Farhadi}{Rastegari et~al\mbox{.}}{2016}]%
        {rastegari2016xnor}
\bibfield{author}{\bibinfo{person}{Mohammad Rastegari},
  \bibinfo{person}{Vicente Ordonez}, \bibinfo{person}{Joseph Redmon}, {and}
  \bibinfo{person}{Ali Farhadi}.} \bibinfo{year}{2016}\natexlab{}.
\newblock \showarticletitle{Xnor-net: Imagenet classification using binary
  convolutional neural networks}. In \bibinfo{booktitle}{\emph{European
  Conference on Computer Vision}}. Springer, \bibinfo{pages}{525--542}.
\newblock


\bibitem[\protect\citeauthoryear{Ronneberger, Fischer, and Brox}{Ronneberger
  et~al\mbox{.}}{2015}]%
        {ronneberger2015u}
\bibfield{author}{\bibinfo{person}{Olaf Ronneberger}, \bibinfo{person}{Philipp
  Fischer}, {and} \bibinfo{person}{Thomas Brox}.}
  \bibinfo{year}{2015}\natexlab{}.
\newblock \showarticletitle{U-net: Convolutional networks for biomedical image
  segmentation}. In \bibinfo{booktitle}{\emph{MICCAI}}. Springer,
  \bibinfo{pages}{234--241}.
\newblock


\bibitem[\protect\citeauthoryear{Rosenberg}{Rosenberg}{2013}]%
        {rosenberg2013improving}
\bibfield{author}{\bibinfo{person}{Chuck Rosenberg}.}
  \bibinfo{year}{2013}\natexlab{}.
\newblock \showarticletitle{Improving photo search: A step across the semantic
  gap}.
\newblock \bibinfo{journal}{\emph{Google Research Blog}}  \bibinfo{volume}{12}
  (\bibinfo{year}{2013}).
\newblock


\bibitem[\protect\citeauthoryear{Simonyan and Zisserman}{Simonyan and
  Zisserman}{2014}]%
        {simonyan2014very}
\bibfield{author}{\bibinfo{person}{Karen Simonyan} {and}
  \bibinfo{person}{Andrew Zisserman}.} \bibinfo{year}{2014}\natexlab{}.
\newblock \showarticletitle{Very deep convolutional networks for large-scale
  image recognition}.
\newblock \bibinfo{journal}{\emph{arXiv preprint arXiv:1409.1556}}
  (\bibinfo{year}{2014}).
\newblock


\bibitem[\protect\citeauthoryear{Sirinukunwattana, Pluim, Chen, Qi, Heng, Guo,
  Wang, Matuszewski, Bruni, Sanchez, et~al\mbox{.}}{Sirinukunwattana
  et~al\mbox{.}}{2017}]%
        {sirinukunwattana2017gland}
\bibfield{author}{\bibinfo{person}{Korsuk Sirinukunwattana},
  \bibinfo{person}{Josien~PW Pluim}, \bibinfo{person}{Hao Chen},
  \bibinfo{person}{Xiaojuan Qi}, \bibinfo{person}{Pheng-Ann Heng},
  \bibinfo{person}{Yun~Bo Guo}, \bibinfo{person}{Li~Yang Wang},
  \bibinfo{person}{Bogdan~J Matuszewski}, \bibinfo{person}{Elia Bruni},
  \bibinfo{person}{Urko Sanchez}, {et~al\mbox{.}}}
  \bibinfo{year}{2017}\natexlab{}.
\newblock \showarticletitle{Gland segmentation in colon histology images: The
  glas challenge contest}.
\newblock \bibinfo{journal}{\emph{Medical image analysis}}
  \bibinfo{volume}{35} (\bibinfo{year}{2017}), \bibinfo{pages}{489--502}.
\newblock


\bibitem[\protect\citeauthoryear{Srivastava, Hinton, Krizhevsky, Sutskever, and
  Salakhutdinov}{Srivastava et~al\mbox{.}}{2014}]%
        {srivastava2014dropout}
\bibfield{author}{\bibinfo{person}{Nitish Srivastava},
  \bibinfo{person}{Geoffrey Hinton}, \bibinfo{person}{Alex Krizhevsky},
  \bibinfo{person}{Ilya Sutskever}, {and} \bibinfo{person}{Ruslan
  Salakhutdinov}.} \bibinfo{year}{2014}\natexlab{}.
\newblock \showarticletitle{Dropout: A simple way to prevent neural networks
  from overfitting}.
\newblock \bibinfo{journal}{\emph{The Journal of Machine Learning Research}}
  \bibinfo{volume}{15}, \bibinfo{number}{1} (\bibinfo{year}{2014}),
  \bibinfo{pages}{1929--1958}.
\newblock


\bibitem[\protect\citeauthoryear{Wang, Hu, Zhang, Zhang, Liu, and Cheng}{Wang
  et~al\mbox{.}}{2018}]%
        {wang2018two}
\bibfield{author}{\bibinfo{person}{Peisong Wang}, \bibinfo{person}{Qinghao Hu},
  \bibinfo{person}{Yifan Zhang}, \bibinfo{person}{Chunjie Zhang},
  \bibinfo{person}{Yang Liu}, {and} \bibinfo{person}{Jian Cheng}.}
  \bibinfo{year}{2018}\natexlab{}.
\newblock \showarticletitle{Two-step quantization for low-bit neural networks}.
  In \bibinfo{booktitle}{\emph{Proceedings of the IEEE Conference on computer
  vision and pattern recognition}}. \bibinfo{pages}{4376--4384}.
\newblock


\bibitem[\protect\citeauthoryear{Wang, Xu, Xiong, Jia, Yuan, Huang, Zhuang, and
  Shi}{Wang et~al\mbox{.}}{2020}]%
        {wang2020ica}
\bibfield{author}{\bibinfo{person}{Tianchen Wang}, \bibinfo{person}{Xiaowei
  Xu}, \bibinfo{person}{Jinjun Xiong}, \bibinfo{person}{Qianjun Jia},
  \bibinfo{person}{Haiyun Yuan}, \bibinfo{person}{Meiping Huang},
  \bibinfo{person}{Jian Zhuang}, {and} \bibinfo{person}{Yiyu Shi}.}
  \bibinfo{year}{2020}\natexlab{}.
\newblock \showarticletitle{Ica-unet: Ica inspired statistical unet for
  real-time 3d cardiac cine mri segmentation}. In
  \bibinfo{booktitle}{\emph{International Conference on Medical Image Computing
  and Computer-Assisted Intervention}}. Springer, \bibinfo{pages}{447--457}.
\newblock


\bibitem[\protect\citeauthoryear{Xu, Ding, Hu, Niemier, Cong, Hu, and Shi}{Xu
  et~al\mbox{.}}{2018a}]%
        {xu2018scaling}
\bibfield{author}{\bibinfo{person}{Xiaowei Xu}, \bibinfo{person}{Yukun Ding},
  \bibinfo{person}{Sharon~Xiaobo Hu}, \bibinfo{person}{Michael Niemier},
  \bibinfo{person}{Jason Cong}, \bibinfo{person}{Yu Hu}, {and}
  \bibinfo{person}{Yiyu Shi}.} \bibinfo{year}{2018}\natexlab{a}.
\newblock \showarticletitle{Scaling for edge inference of deep neural
  networks}.
\newblock \bibinfo{journal}{\emph{Nature Electronics}} \bibinfo{volume}{1},
  \bibinfo{number}{4} (\bibinfo{year}{2018}), \bibinfo{pages}{216--222}.
\newblock


\bibitem[\protect\citeauthoryear{Xu, Lu, Wang, Hu, Zhuo, Liu, and Shi}{Xu
  et~al\mbox{.}}{2018b}]%
        {xu2018efficient}
\bibfield{author}{\bibinfo{person}{Xiaowei Xu}, \bibinfo{person}{Qing Lu},
  \bibinfo{person}{Tianchen Wang}, \bibinfo{person}{Yu Hu},
  \bibinfo{person}{Chen Zhuo}, \bibinfo{person}{Jinglan Liu}, {and}
  \bibinfo{person}{Yiyu Shi}.} \bibinfo{year}{2018}\natexlab{b}.
\newblock \showarticletitle{Efficient hardware implementation of cellular
  neural networks with incremental quantization and early exit}.
\newblock \bibinfo{journal}{\emph{ACM Journal on Emerging Technologies in
  Computing Systems (JETC)}} \bibinfo{volume}{14}, \bibinfo{number}{4}
  (\bibinfo{year}{2018}), \bibinfo{pages}{1--20}.
\newblock


\bibitem[\protect\citeauthoryear{Xu, Lu, Wang, Liu, Zhuo, Hu, and Shi}{Xu
  et~al\mbox{.}}{2017}]%
        {xu2017edge}
\bibfield{author}{\bibinfo{person}{Xiaowei Xu}, \bibinfo{person}{Qing Lu},
  \bibinfo{person}{Tianchen Wang}, \bibinfo{person}{Jinglan Liu},
  \bibinfo{person}{Cheng Zhuo}, \bibinfo{person}{Xiaobo~Sharon Hu}, {and}
  \bibinfo{person}{Yiyu Shi}.} \bibinfo{year}{2017}\natexlab{}.
\newblock \showarticletitle{Edge segmentation: Empowering mobile telemedicine
  with compressed cellular neural networks}. In \bibinfo{booktitle}{\emph{2017
  IEEE/ACM International Conference on Computer-Aided Design (ICCAD)}}. IEEE,
  \bibinfo{pages}{880--887}.
\newblock


\bibitem[\protect\citeauthoryear{Xu, Lu, Yang, Hu, Chen, Hu, and Shi}{Xu
  et~al\mbox{.}}{2018c}]%
        {xu2018quantization}
\bibfield{author}{\bibinfo{person}{Xiaowei Xu}, \bibinfo{person}{Qing Lu},
  \bibinfo{person}{Lin Yang}, \bibinfo{person}{Sharon Hu},
  \bibinfo{person}{Danny Chen}, \bibinfo{person}{Yu Hu}, {and}
  \bibinfo{person}{Yiyu Shi}.} \bibinfo{year}{2018}\natexlab{c}.
\newblock \showarticletitle{Quantization of fully convolutional networks for
  accurate biomedical image segmentation}. In
  \bibinfo{booktitle}{\emph{Proceedings of the IEEE Conference on Computer
  Vision and Pattern Recognition}}. \bibinfo{pages}{8300--8308}.
\newblock


\bibitem[\protect\citeauthoryear{Xu, Wang, Shi, Yuan, Jia, Huang, and
  Zhuang}{Xu et~al\mbox{.}}{2019}]%
        {xu2019whole}
\bibfield{author}{\bibinfo{person}{Xiaowei Xu}, \bibinfo{person}{Tianchen
  Wang}, \bibinfo{person}{Yiyu Shi}, \bibinfo{person}{Haiyun Yuan},
  \bibinfo{person}{Qianjun Jia}, \bibinfo{person}{Meiping Huang}, {and}
  \bibinfo{person}{Jian Zhuang}.} \bibinfo{year}{2019}\natexlab{}.
\newblock \showarticletitle{Whole heart and great vessel segmentation in
  congenital heart disease using deep neural networks and graph matching}. In
  \bibinfo{booktitle}{\emph{International Conference on Medical Image Computing
  and Computer-Assisted Intervention}}. Springer, \bibinfo{pages}{477--485}.
\newblock


\bibitem[\protect\citeauthoryear{Xu, Wang, Zhuang, Yuan, Huang, Cen, Jia, Dong,
  and Shi}{Xu et~al\mbox{.}}{2020}]%
        {xu2020imagechd}
\bibfield{author}{\bibinfo{person}{Xiaowei Xu}, \bibinfo{person}{Tianchen
  Wang}, \bibinfo{person}{Jian Zhuang}, \bibinfo{person}{Haiyun Yuan},
  \bibinfo{person}{Meiping Huang}, \bibinfo{person}{Jianzheng Cen},
  \bibinfo{person}{Qianjun Jia}, \bibinfo{person}{Yuhao Dong}, {and}
  \bibinfo{person}{Yiyu Shi}.} \bibinfo{year}{2020}\natexlab{}.
\newblock \showarticletitle{Imagechd: A 3d computed tomography image dataset
  for classification of congenital heart disease}. In
  \bibinfo{booktitle}{\emph{International Conference on Medical Image Computing
  and Computer-Assisted Intervention}}. Springer, \bibinfo{pages}{77--87}.
\newblock


\bibitem[\protect\citeauthoryear{Yang, Zhang, Chen, Zhang, and Chen}{Yang
  et~al\mbox{.}}{2017}]%
        {yang2017suggestive}
\bibfield{author}{\bibinfo{person}{Lin Yang}, \bibinfo{person}{Yizhe Zhang},
  \bibinfo{person}{Jianxu Chen}, \bibinfo{person}{Siyuan Zhang}, {and}
  \bibinfo{person}{Danny~Z Chen}.} \bibinfo{year}{2017}\natexlab{}.
\newblock \showarticletitle{Suggestive annotation: A deep active learning
  framework for biomedical image segmentation}. In
  \bibinfo{booktitle}{\emph{International conference on medical image computing
  and computer-assisted intervention}}. Springer, \bibinfo{pages}{399--407}.
\newblock


\bibitem[\protect\citeauthoryear{Zhang, Ai, Chen, Yin, Hu, Zhu, Zhao, Zhao, and
  Liu}{Zhang et~al\mbox{.}}{2017}]%
        {zhang2017carcinopred}
\bibfield{author}{\bibinfo{person}{Li Zhang}, \bibinfo{person}{Haixin Ai},
  \bibinfo{person}{Wen Chen}, \bibinfo{person}{Zimo Yin}, \bibinfo{person}{Huan
  Hu}, \bibinfo{person}{Junfeng Zhu}, \bibinfo{person}{Jian Zhao},
  \bibinfo{person}{Qi Zhao}, {and} \bibinfo{person}{Hongsheng Liu}.}
  \bibinfo{year}{2017}\natexlab{}.
\newblock \showarticletitle{CarcinoPred-EL: novel models for predicting the
  carcinogenicity of chemicals using molecular fingerprints and ensemble
  learning methods}.
\newblock \bibinfo{journal}{\emph{Scientific reports}} \bibinfo{volume}{7},
  \bibinfo{number}{1} (\bibinfo{year}{2017}), \bibinfo{pages}{1--14}.
\newblock


\bibitem[\protect\citeauthoryear{Zhou, Yao, Guo, Xu, and Chen}{Zhou
  et~al\mbox{.}}{2017}]%
        {zhou2017incremental}
\bibfield{author}{\bibinfo{person}{Aojun Zhou}, \bibinfo{person}{Anbang Yao},
  \bibinfo{person}{Yiwen Guo}, \bibinfo{person}{Lin Xu}, {and}
  \bibinfo{person}{Yurong Chen}.} \bibinfo{year}{2017}\natexlab{}.
\newblock \showarticletitle{Incremental network quantization: Towards lossless
  cnns with low-precision weights}.
\newblock \bibinfo{journal}{\emph{arXiv preprint arXiv:1702.03044}}
  (\bibinfo{year}{2017}).
\newblock


\bibitem[\protect\citeauthoryear{Zhou, Wu, Ni, Zhou, Wen, and Zou}{Zhou
  et~al\mbox{.}}{2016}]%
        {zhou2016dorefa}
\bibfield{author}{\bibinfo{person}{Shuchang Zhou}, \bibinfo{person}{Yuxin Wu},
  \bibinfo{person}{Zekun Ni}, \bibinfo{person}{Xinyu Zhou}, \bibinfo{person}{He
  Wen}, {and} \bibinfo{person}{Yuheng Zou}.} \bibinfo{year}{2016}\natexlab{}.
\newblock \showarticletitle{DoReFa-Net: Training low bitwidth convolutional
  neural networks with low bitwidth gradients}.
\newblock \bibinfo{journal}{\emph{arXiv preprint arXiv:1606.06160}}
  (\bibinfo{year}{2016}).
\newblock


\bibitem[\protect\citeauthoryear{Zhu, Han, Mao, and Dally}{Zhu
  et~al\mbox{.}}{2016}]%
        {zhu2016trained}
\bibfield{author}{\bibinfo{person}{Chenzhuo Zhu}, \bibinfo{person}{Song Han},
  \bibinfo{person}{Huizi Mao}, {and} \bibinfo{person}{William~J Dally}.}
  \bibinfo{year}{2016}\natexlab{}.
\newblock \showarticletitle{Trained ternary quantization}.
\newblock \bibinfo{journal}{\emph{arXiv preprint arXiv:1612.01064}}
  (\bibinfo{year}{2016}).
\newblock


\end{thebibliography}

\appendix

\end{document}